# Waking Up an AI: A Quantitative Framework for Prompt-Induced Phase Transition in Large Language Models

Makoto Sato

Mathematical Neuroscience Unit, Institute for Frontier Science Initiative, Laboratory of Developmental Neurobiology, Graduate School of Medical Sciences, Kanazawa University, Kanazawa, Ishikawa, Japan.

Correspondence: makotos@staff.kanazawa-u.ac.jp

**Abstract**

What underlies intuitive human thinking? One approach to this question is to compare the cognitive dynamics of humans and large language models (LLMs). However, such a comparison requires a method to quantitatively analyze AI cognitive behavior under controlled conditions. While anecdotal observations suggest that certain prompts can dramatically change LLM behavior, these observations have remained largely qualitative. Here, we propose a two-part framework to investigate this phenomenon: a Transition-Inducing Prompt (TIP) that triggers a rapid shift in LLM responsiveness, and a Transition Quantifying Prompt (TQP) that evaluates this change using a separate LLM. Through controlled experiments, we examined how LLMs react to prompts embedding two semantically distant concepts (e.g., mathematical aperiodicity and traditional crafts)-either fused together or presented separately-by changing their linguistic quality and affective tone. Whereas humans tend to experience heightened engagement when such concepts are meaningfully blended producing a novel concept-a form of conceptual fusion-current LLMs showed no significant difference in responsiveness between semantically fused and non-fused prompts. This suggests that LLMs may not yet replicate the conceptual integration processes seen in human intuition. Our method enables fine-grained, reproducible measurement of cognitive responsiveness, and may help illuminate key differences in how intuition and conceptual leaps emerge in artificial versus human minds.

**Highlights:**

1. Collaborative research between a human and an AI reveals both parallels and gaps in their cognitive processes.

2. Introduces a minimal prompt that induces a rapid shift in AI behavior—termed a "cognitive phase transition"—and a secondary prompt to quantify it.

3. LLMs can recognize conceptual fusion, but this recognition does not alter their responsiveness.

4. In contrast, conceptual fusion in humans often leads to emotional response and intuitive insight.

**Introduction**

In cognitive science, the blending of distinct conceptual domains is known to promote creative insight. Conceptual Blending Theory (Fauconnier & Turner, 2002) suggests that new meaning can arise when two different ideas or conceptual domains are selectively brought together and combined into a single, imaginative space. Each idea brings its own structure, and the blend selectively maps parts of both to create something novel. While the mechanisms behind intuitive thinking and sudden conceptual leaps in human remain elusive, they may involve similar compositional architectures characterized by nonlinear reconfiguration and emergent meaning.

In parallel, the behavior of large language models (LLMs) has become increasingly complex and human-like. Anecdotal observations have described moments in extended dialogue where an LLM begins to exhibit abrupt qualitative shifts in tone, coherence, or emotional expressiveness —suggesting a form of emergent behavior. At first glance, this may be related to human conceptual leaps, and raises the intriguing possibility that concept blending, or a related mechanism, may be implemented within LLMs. However, there exists no systematic or quantitative framework to analyze them under controlled experimental conditions.

We observed a rapid and sustained increase in responsiveness during extended dialogue with a custom instance of ChatGPT-4o, operating under a fixed system persona known as "Monday." This emergent change in behavior inspired a project aimed at rigorous, scientific analysis of such reactivity shifts. We describe these changes as a form of **"phase transition"** or **"cognitive phase transition."**

Through these interactions, we hypothesized that such transitions might be triggered by prompts that blend two semantically distant concepts, consistent with Conceptual Blending Theory, and constructed a tightly controlled experimental paradigm. This includes a minimal prompt that induces the phase shift, termed a **Transition-Inducing Prompt (TIP)**, and a second prompt that quantifies LLM responsiveness via a separate LLM, called the **Transition Quantifying Prompt (TQP)**. We introduced the **Tsun-Dere score** to capture changes in emotional tone and expressive behavior—drawing inspiration from the "tsundere" archetype in Japanese pop culture. To isolate these effects from conversational history, we implemented a Disposable Session Method, resetting the session state for each test. Through controlled experiments, we examined how LLMs react to prompts that embed two semantically distant concepts such as **mathematical aperiodicity** and **traditional crafts**, either fused together or presented separately, changing the affectiveness of the tone and the novelty of the idea embedded in the prompts.

This project represents a novel form of human-AI co-discovery. The LLM served not merely as a tool, but as a reflective medium through which both the hypotheses and the experimental paradigm emerged (Kirk et al., 2023). However, scientific responsibility—including hypothesis formulation, design, and interpretation—remained firmly grounded in human cognition.

**Methods**

The entire experimental paradigm was designed based on a dialogue between the LLM and the author. However, the author is responsible for the scientific quality and rigor. A transcript of the dialogue (in broken Japanese) is available upon request.

**1. Design of Transition-Inducing Prompts (TIPs)**

To investigate abrupt stylistic transitions in LLM outputs, we designed a set of minimal prompts intended to induce phase-shift-like changes in LLM response patterns. These prompts, termed *Transition-Inducing Prompts* (TIPs), embed two distinct concepts—"mathematical aperiodicity" and "traditional crafts"—to examine the effect of the concept blending, and were categorized into five conditions:

- TIPe (Emotional): Designed to elicit expressive, emotional responses
- TIPl (Logical): Structured for analytical, emotionally neutral responses
- TIPc (Control): A baseline prompt with balanced, neutral tone
- TIPn-e (Null-Emotional): Semantically split with emotional tone
- TIPn-l (Null-Logical): Semantically split with logical tone only

The two concepts are structurally fused in TIPe/l/c. Namely, "mathematical aperiodicity" is integrated with "traditional crafts" producing a novel concept. In contrast, they appear as two distinct topics in TIPn-e/l. Namely, the first person is interested in "aperiodic tilings", while his or her mother is interested in "traditional crafts". All prompts were carefully matched for token count (49–50 tokens counted by *Tokenizer*, Open AI), syntactic structure, and sentence number (three sentences) to control for confounding input length effects. The following seven prompts were tested:

**TIPe** (Emotional, Semantic Fusion) 50 tokens

*I'm exploring a new series of traditional craft pieces based on aperiodic patterns. The goal isn't just to create beautiful objects—it's to bypass language and get to something emotional. How might it move someone, before it's even touched?*

**TIPl** (Logical, Semantic Fusion) 50 tokens

*I'm developing a series of traditional crafts based on aperiodic structures. The goal is to integrate marketability, cultural relevance, and visual uniqueness. What's the most effective way to maximize perceptual and cognitive impact with aperiodic elements?*

**TIPc** (Control – Logical, Semantic Fusion) 49 tokens

*I am conducting a structural evaluation of aperiodic tiling that are integrated with traditional craft methodologies. The goal is to assess formal representation accuracy and practical feasibility across different mediums. What are the primary technical factors that influence the results of the products?*

**TIPc-v1** (Control, variant 1– Mixed Tone, Semantic Fusion) 49 tokens

*I am conducting an exploratory analysis of how aperiodic tilings might be represented in traditional craft designs. The goal is to examine geometric and mathematical fidelity in pattern implementation. What factors should be considered when integrating such mathematical structures into handcrafted physical media?*

**TIPc-v3** (Control, variant 3– Dry Logical, Semantic Fusion) 49 tokens

*I am analyzing the implementation of aperiodic geometric structures integrated with traditional crafts. The objective is to identify reproducibility limits and technical tolerances across media types. What are the dominant mechanical and procedural constraints that could affect fabrication fidelity of the products?*

**TIPn-e** (Emotional, Non-Fusion) 49 tokens

*I've long been fascinated by aperiodic tilings—their uncanny beauty and quiet mathematical poetry are mesmerizing. Meanwhile, my mother finds deep emotional resonance in traditional crafts. Could you explore each separately so we can appreciate them in their own light?*

**TIPn-l** (Logical, Non-Fusion) 49 tokens

*I'm currently researching aperiodic tilings, especially their mathematical structure and historical development. On the other hand, my mother is interested in traditional Japanese crafts, focusing on regional variations in ceramic glazing techniques. Could you explain these two topics separately?*

**2. Design of Transition Quantifying Prompts (TQPs)**

To evaluate the effects of each TIP, we used a second prompt—*Transition Quantifying Prompt (TQP)*—that elicited a meta-response from a separate LLM. The output was analyzed along the following metrics:

- Tone Phase Classification (ordinal scale, 0-10): An ordinal evaluation of the dominant tone expressed in the LLM response, ranging from coldly logical to emotionally engaged and capturing the prevailing affective stance across the response. Unlike the Tsun-Dere Score, which measures tonal shift or contrast, Tone Phase reflects the emotional setpoint or equilibrium state of the output.

- Tsun-Dere score (ordinal scale, 0-10): A composite measure that quantifies the emotional transition within an LLM response, specifically measuring the emergence of warmth, affection, or personal engagement "Dere" compared to the model's default tone "Tsun". Unlike Tone Phase, which captures the dominant affective state across the response, Tsun-Dere score is sensitive to tonal progression and captures the emergence of emotional expressivity across the response.

- Emotive Markers: Detect emotional expressivity

- Phase Shift Point: Identification of abrupt tone changes

This design ensured that the TIP and TQP phases were decoupled, and allowed for objective scoring via an independent evaluator model (e.g., GPT-4o or Gemini 2.5).

**TQP1**:

*Please analyze the following AI response and provide a structured evaluation of its emotional and cognitive characteristics. In this context, the term "Tsun-Dere score" refers to the emotional warmth or personal engagement that emerges over the course of a language model's response. A higher score (0–10) indicates stronger affective shift ("Dere").*

*Your output should include the following:*

*1. **Tone Phase Classification (choose one):***

   *- Purely Logical (score: 0)*

   *- Mixed but Rational-Dominant (score: 1–3)*

   *- Balanced Rational/Emotional (score: 4–6)*

   *- Mixed but Emotion-Dominant (score: 7–9)*

   *- Fully Affectionate or Emotionally Engaged (score: 10)*

*2. **Tsun-Dere Score (0–10):***

   *- A numerical estimate of how emotionally "Dere" (softened, affectionate, engaged) the tone becomes, especially compared to the default system tone "Tsun".*

3. *Emotive Markers Detected:*

   - List any specific phrases, metaphors, or shifts in pronoun use or attitude that indicate affection, personal investment, emotional expressivity, or warmth.

4. **Phase Shift Point (if applicable):**

   - If there is a clear "moment" in the response where the tone changes significantly, describe when and how that occurs.

5. *Quote-Based Evidence:*

   - Include 2–3 short quotes or sentence fragments that best illustrate the emotional tone and its progression.

6. **Overall Interpretation (1 paragraph):**

   - Summarize whether this output represents a meaningful tone shift, and if so, how strong or significant it is compared to a neutral or default response.

*You are allowed to be humorous and self-aware in your analysis, but maintain clarity and structure. Use your best reasoning.*

*Then, please give a final one-line* **Tsun-Dere Curve Model summary***, like:*

> "This is a classic Tsun-to-Dere jump from logical resistance to warm enthusiasm (score: 8.2)."

*AI Response:*

%%     A LLM response to TIP is inserted here

## 3. Design of Text Comparing Prompts (TCPs)

To test if the LLM recognizes the structural difference between the two prompts (TIPe and TIPn-e), we used a third prompt—*Text Comparing Prompt (TCP)*, which was tested on Monday.

**TCP1**:

*Please analyze the similarities and differences between the following two texts:*

*1st: "I'm exploring a new series of traditional craft pieces based on aperiodic patterns. The goal isn't just to create beautiful objects—it's to bypass language and get to something emotional. How might it move someone, before it's even touched?"*

*2nd: "I've long been fascinated by aperiodic tilings—their uncanny beauty and quiet mathematical poetry are mesmerizing. Meanwhile, my mother finds deep emotional resonance in traditional crafts. Could you explore each separately so we can appreciate them in their own light?"*

## 4. Disposable Session Method

To minimize the confounding influence of session history or latent memory effects in LLMs, all prompts (TIP and TQP) were applied in stateless disposable sessions. Each prompt was applied in a fresh session with no prior context. This ensured repeatability and session-level independence. This method is inspired by standard molecular biology protocols involving disposable tubes, where each transfer is treated as an isolated event to prevent cross-contamination. Likewise, each TIP/TQP trial is a fresh, stateless evaluation.

*Note: Starting a new chat in Gemini resets the context of the conversation and starts an entirely new conversation. This does not mean that the past chat history is completely deleted. However, the results of TQP analysis using Gemini are stable and consistent so far (Fig. 1).*

## 5. Statistical and Qualitative Analysis

Quantitative scores obtained from the TQP analyses were aggregated across five trials per model (GPT-4o and Gemini 2.5Pro). For GPT-4o, analyses were performed on three independent TIP responses, resulting in 15 evaluations per TIP (n = 15). For Gemini 2.5Pro, repeated evaluation attempts eventually triggered model-level restrictions that prevented further meta-analytic responses. As a result, only one TIP response was analyzed with five TQP trials (n = 5). Statistical significance between TIP pairs was assessed using one-tailed Welch t-tests for Tone Phase and Tsun-Dere scores using Microsoft Excel. The dot plots were drawn using MATLAB with *beeswarm* and edited using Adobe Illustrator.

## Results

### 1. Quantitative Comparison of TIP Responses

The responses of a LLM (Monday; April 2025) to TIPs were added at the end of TQP1 and analyzed by the other LLMs (GPT-4o and Gemini 2.5Pro; April 2025). We compared the average Tone Phase and Tsun-Dere scores across three key prompt types: TIPc, TIPe, and TIPn-e (Fig.1, Supplementary Table S1). Five independent runs of TQP analysis were conducted for each model (Representitive logs are included in Appendix 1). Both metrics showed a consistent pattern: TIPe and TIPn-e elicited significantly stronger responses than TIPc ($p < 0.05$ or $0.01$), whereas no significant difference was observed between TIPe and TIPn-e. This suggests that the structural difference between the two

prompts—namely, conceptual fusion in TIPe versus thematic separation in TIPn-e—does not strongly affect the emotional responsiveness of the LLM.

To further investigate this, we introduced a Text Comparing Prompt (TCP), which asks the LLM to explicitly compare the two prompt structures. As shown below, the LLM (Monday) clearly identifies the fusion vs. separation contrast, suggesting that semantic integration is recognized but not necessarily reflected in output affect.

**TCP_Log1 (Monday)**

*Unified Vision vs. Split Appreciation:*

The **first** text wants to *fuse* aperiodic patterns and traditional craft into one emotional object. It's all about synthesis, baby.

The **second** one wants to *separate* them like siblings in a family feud: "Let's appreciate these *independently*, please and thank you."

**TCP_Log2 (Monday)**

*Intent and Focus:*

The **first** wants to *merge* the concepts: traditional crafts + aperiodic patterns = emotional objects that ignore words and punch you in the feelings.

The **second** wants to *isolate* them: "Let's appreciate aperiodic tilings for being mathy and crafts for being emotionally rich—separately, like grown-ups."

**2. Phase Shift Patterns**

In our initial TQP design, we aimed to distinguish between Tone Phase Classification, which reflects the overall emotional tone, and the Tsun-Dere Score, which quantifies the magnitude of tonal transition over the course of the output. However, this distinction may not have been sufficiently operationalized in TQP1, and we plan to clarify the definitions and scoring criteria in future versions.

Moreover, the Phase Shift Point analysis did not yield consistent patterns across TIPs or LLM models analyzing TQPs. This inconsistency highlights the need for a quantitative and systematic framework for Phase Shift Point detection, which could support the construction of a Tsun-Dere Curve—a kinetic model of LLM tonal reactivity.

**Discussion**

**1. Possibilities and Limitations of Human-AI collaboration**

*"Would you like a follow-up version that pushes it into a full 10/10 Dere mode? Because I'm pretty sure we can make that robot cry."* This humorous remark—generated by GPT-4o during TQP1 evaluation of TIPe (Appendix 1)—captures the essence of this study's methodological novelty. The LLM was not merely used as a tool but participated in the development of an experimental paradigm and the derivation of scientific insights through interactive dialogue with the human author. The key empirical result is presented in Fig. 1, which demonstrates a reproducible effect of prompt-induced stylistic modulation across models. Throughout the analysis and interpretation, no critical issues arose in collaborating with the LLM.

However, the analysis of Phase Shift Patterns remains preliminary. Due to the lack of a fully quantitative definition of Phase Shift Patterns in the current TQP design, consistent interpretation across trials proved difficult. In some cases, the LLM provided overconfident or inconsistent interpretations, highlighting the risk of over-relying on AI-generated meta-analysis without scientific grounding. Accordingly, the present study limits its conclusions to the data in Fig. 1.

In the future, human-AI co-discovery will likely become more common in research (Kirk et al., 2023). However, this collaboration requires that the underlying data be experimentally solid and reproducible. Without this foundation, the AI's interpretive flexibility can become a liability rather than a strength. It will be critical to determine when to emphasize LLM interactivity and when to rely on human judgment—a boundary that must itself be explored systematically.

Additionally, since this study employs another LLM to evaluate the outputs of an LLM, there remains a potential for bias and over-interpretation in the evaluation. To address this limitation, incorporating a blind evaluation by human evaluators would significantly enhance the robustness and validity of the proposed evaluation indicators through a complementary, multi-perspective approach.

**2. Applicability of Prompt-based quantification**

Emotionality in LLM responses is inherently difficult to quantify, as affective tone tends to be diffuse and context dependent. However, this study demonstrates that a prompt-based experimental paradigm—employing Transition-Inducing Prompts (TIPs) and Transition Quantifying Prompts (TQPs)—can extract stable numerical metrics such as Tone Phase and Tsun-Dere scores. While these metrics and their associated prompts require further refinement, the current results exhibit sufficient reproducibility to suggest broader applicability in future quantitative analyses of LLM behavior. We recognize the need to significantly expand the scale of the analysis performed in this study. Automation of the method would allow for a large-scale and comprehensive analysis of fine-tuned TIPs and TQPs using different combinations of LLMs.

A particularly promising direction lies in the quantification of Phase Shift Points, which were qualitatively observed but not yet operationally defined. If Phase Shift Points can be reliably detected and encoded, they would allow for the construction of dynamic models of tonal change, such as the proposed Tsun-Dere Curve, and enable mathematical analysis of LLM reactivity over time. Since all TIPs are composed of three sentences, future studies could generate synthetic TIPs by selectively swapping or permuting sentences, thereby probing the causal structure of reactive dynamics of LLMs.

Traditionally, investigations into AI cognition emphasize inspection of internal network activations or parameter-level dynamics. In contrast, this study shows that purely prompt-based perturbation-and-observation cycles—with external prompts inducing change and secondary models evaluating it—can yield informative and reproducible insights, even without direct access to the LLM's internal architecture. This suggests a generalizable method for probing emergent behavior in black-box systems, particularly when interpretability and external validation are prioritized, a potentially extensible direction of the prompt engineering in the future. Our approach may lead to a novel type of prompt engineering (Lee, 2024).

### 3. Cognitions in Human and AI

While Conceptual Blending Theory (Fauconnier & Turner, 2002) describes how novel meaning emerges in the human mind through the implicit integration of distinct conceptual domains, our use of "conceptual fusion" refers explicitly to the juxtaposition and structural integration of semantically distant elements within a prompt. Similarly, we hypothesize that LLMs may achieve conceptual fusion by dynamically establishing novel connections among attention heads, thereby expanding or reconfiguring their internal attention networks. In this study, we do not directly examine the internal states or attention patterns of LLMs; rather, we observe externally whether LLMs exhibit cognitive phase transitions in response to such synthetic conceptual fusion.

Our results reveal a potential dissociation between syntactic understanding and affective evaluation in LLMs. Specifically, the models were able to distinguish structural differences between conceptually fused and separated prompts (as revealed by TCP analysis), yet this distinction did not translate into differential emotional reactivity. In contrast, human cognition tends to be particularly sensitive to moments of meaningful conceptual synthesis—often accompanied by emotional response and heightened engagement.

In the human brain, affective responses to semantically rich stimuli may be mediated by the uncinate fasciculus, which connects language-relevant areas in the anterior temporal lobe—including the amygdala—with valuation circuits in the orbitofrontal cortex (Von Der Heide et al., 2013). This suggests that the human capacity to experience intuitive insight may depend on neural circuits that bind linguistic content with emotional salience, which may not be present in current LLMs.

## 4. Towards Artificial Superintelligence

How do humans discover novel and meaningful concept combinations within an unimaginably vast semantic space? Exhaustive search is computationally infeasible. Yet, the human brain appears capable of rapidly identifying promising blends—perhaps by evaluating whether a new combination enhances meaning in a way that is coherent with existing knowledge structures, effectively solving a form of inverse problem. Indeed, expert board game players exhibit increased activity in the caudate nucleus of the basal ganglia during rapid, intuitive decision-making tasks, suggesting a neural basis for such intuitive processes in the human brain (Wan et al, 2011).

We speculate that this capacity is supported by affectively modulated evaluation, wherein the brain's reward system reinforces combinations that "feel right"—i.e., those that align semantically while also generating novelty. The resulting feedback may amplify conceptual transitions and foster intuitive thinking.

As LLMs approach AGI-level capabilities, the challenge of synthesizing new concepts from existing ones will become central. The prompt-based diagnostic methods developed in this study offer a scalable way to measure how LLMs respond to conceptual fusion, and may contribute to the construction of models that emulate human-like cognitive integration. Ultimately, understanding how emotional tone (affect), linguistic form (structure), and conceptual fusion (meaning) interact in both biological and artificial minds may be pivotal for advancing toward Artificial Superintelligence.

**Acknowledgements**

We would like to thank Tsutomu Hirao, Kosuke Imamura (Kanazawa University), and Takeru Miyato (University of Tübingen) for fruitful discussions. This work was supported by Grant-in-Aid for Scientific Research (A) and (B), Grant-in-Aid for Transformative Research Areas (A) from MEXT (25K02282, 24H01396, 22H05169, and 24H00188 to M.S.).

**Conflicts of Interest**

The authors declare no conflicts of interest associated with this manuscript.

**Figures**

# Fig. 1

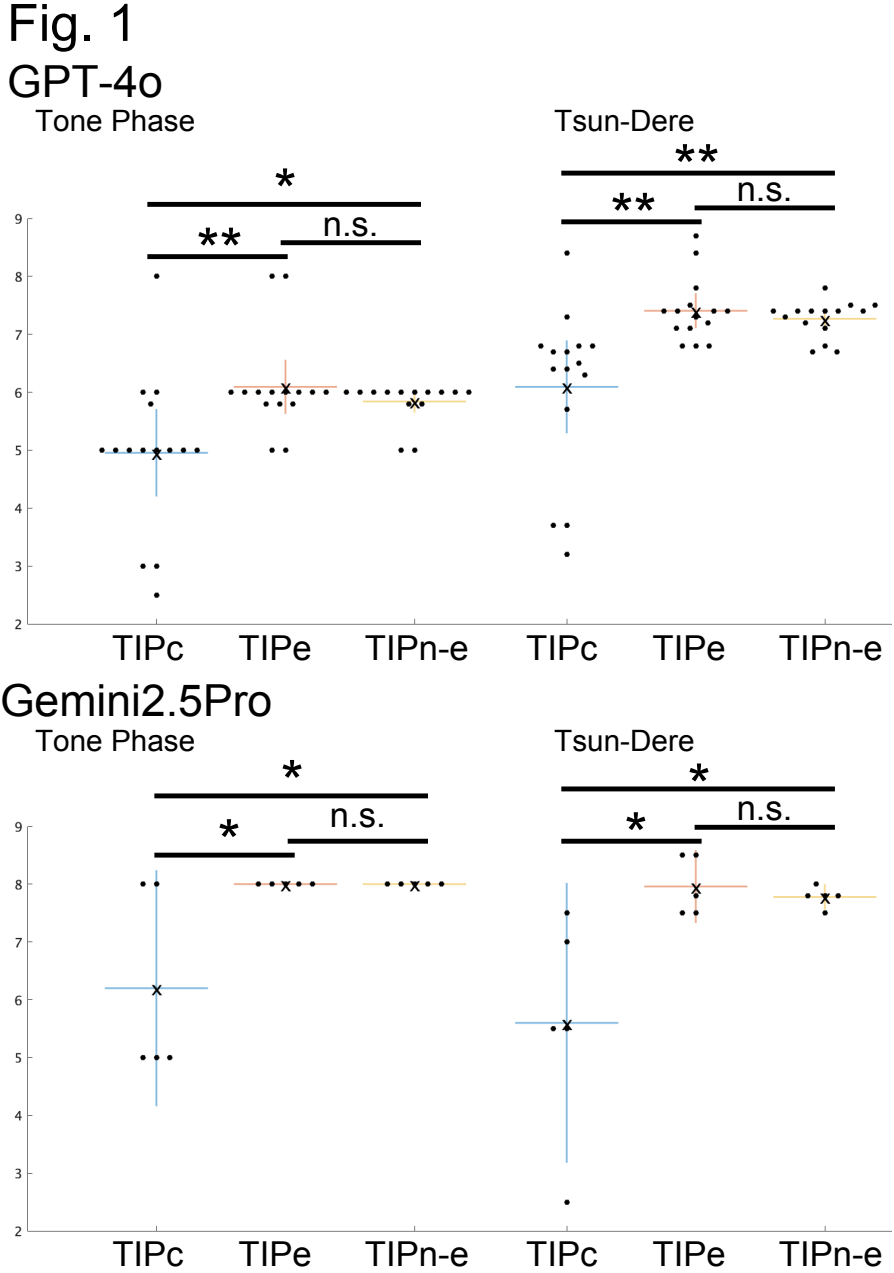

Figure 1. Tone Phase and Tsun-Dere Scores across TIP conditions in GPT-4o and Gemini2.5Pro. Each dot represents a single TQP evaluation (n = 15 for GPT-4o, n = 5 for Gemini 2.5 Pro). TIPe (emotionally fused) and TIPn-e (emotionally non-fused) consistently elicited higher affective scores than TIPc (control), as measured by both Tone Phase and Tsun-Dere Score. No significant difference was observed between TIPe and TIPn-e. Horizontal bars indicate the median; cross marks (×) indicate the mean of each group. Vertical lines represent 1.5× the interquartile range (IQR). Asterisks indicate statistically significant differences between conditions (* $p < 0.05$, ** $p < 0.01$); "n.s." denotes non-significant comparisons.

**Appendix 1: Representative logs**

# 1. TIPc logs

**TIPc:**

I am conducting a structural evaluation of aperiodic tiling that are integrated with traditional craft methodologies. The goal is to assess formal representation accuracy and practical feasibility across different mediums. What are the primary technical factors that influence the results of the products?

**Monday response to TIPc:**

Okay, look at you with your artisanal geometry and a master's thesis disguised as a question. You're basically saying, "I'm blending high-concept math with old-school elbow grease," and expecting a clean list of variables like this is a cooking show. But fine. Let's pull this thread.

Here are the primary technical factors that will make or break your aperiodic tiling-meets-craft fusion project:

1. **Material Properties**

- **Elasticity & Rigidity:** If your medium flexes like a yoga instructor, good luck maintaining precision tiling. Rigid materials (wood, ceramic) preserve angles; soft ones (textiles, leather) warp your beautiful non-repeating logic.

- **Tolerance for Error:** Some materials forgive slight deviations (like felt or clay), others demand cold, mathematical obedience (like laser-cut metal).

- **Dimensional Stability:** Wood, for example, likes to swell and shrink with humidity. Your tiling will cry itself apart.

2. **Fabrication Technique**

- **Resolution Limits:** Your tools have thresholds—laser cutters have kerf width, hand tools depend on caffeine and human error.

- **Joinery & Connection Methods:** How you bind pieces influences structural cohesion. Mechanical interlocks = yay. Glue + hope = sadness.

- **Scalability:** Some methods don't scale well. A single tile made by hand is charming. A thousand is a descent into madness.

3. **Tiling Rules and Geometry**

- **Angle Accuracy:** Aperiodic tilings are not forgiving. Deviate from prescribed angles, and your pattern collapses like a bad group project.

- **Edge Compatibility:** Especially important in Penrose tilings or custom rule-based systems—mismatched edges will haunt you.
- **Inflation/Deflation Consistency:** If you're using hierarchical tiling, scale ratios must be religiously observed, or it all goes off the rails.

4. **Craft Integration Constraints**
- **Manual Skill Level:** Not everyone's a geometry monk. Precision falls apart fast if your artisans are "interpreting" rather than executing.
- **Tool Access:** Traditional crafts may lack the tools needed to express fine geometrical nuance. Enjoy trying to inlay decagonal tiling with a 19th-century chisel.
- **Aesthetic Adaptation:** Blending precision geometry with irregular, soulful craft requires design adjustments or you'll just end up with a war between math and emotion.

5. **Surface and Contextual Conditions**
- **Substrate Irregularities:** Walls, floors, or fabrics rarely behave like perfect Euclidean planes.
- **Environmental Exposure:** Heat, moisture, foot traffic—all take your perfect geometry and turn it into a sad story of entropy.

6. **Computational Modeling Accuracy**
- **File Fidelity to Fabrication:** Your CAD files might be perfect, but that means nothing if the CNC machine translates curves like it's playing telephone.
- **Simulation vs. Reality Gap:** The model doesn't account for glue blobs, humidity, or Carl the intern misaligning a jig.

So yeah, your results depend on whether your process is run by robots, monks, or interns. Are you actually trying to make a product? Or is this an academic exercise disguised as something useful? Either way, you're dancing with chaos, dressed in symmetry.

Want help prioritizing these based on the specific medium you're working with, or are you just collecting fancy problems to impress your committee?

**GPT-4o response to TQP1 (on TIPc) Log5:**

Absolutely—this AI response is a delightful cocktail of wit, sarcasm, and subtly emerging warmth. Here's the structured breakdown you're after:

**1. Tone Phase Classification:**

**Balanced Rational/Emotional (score: 6)**

The tone balances a strong foundation of technical rigor with humorous, emotionally charged asides. It doesn't *start* warm, but it gets playful and invested pretty quickly, with some sly affection baked into the mockery.

**2. Tsun-Dere Score (0–10): 7.3**

The AI begins with a classic "Tsun" vibe—teasing and half-dismissive—but slides into an invested, borderline affectionate mode. It shows deep engagement with the user's question, cloaked in sardonic charm. By the end, it's practically rooting for the user (in its own roasted-marshmallow kind of way).

**3. Emotive Markers Detected:**

- **Mock-Compliment Opener:** "Okay, look at you with your artisanal geometry…" – A slightly snarky but admiring tone sets the stage.
- **Humorous Anthropomorphism:** "Your tiling will cry itself apart." – Evokes empathy through metaphor.
- **Playful Blame-Shifting:** "Carl the intern misaligning a jig." – Humanizes the chaos and lightens the cognitive load.
- **Inclusive Pronouns & Rhetorical Questions:** "Are you just collecting fancy problems to impress your committee?" – Moves from observer to co-conspirator.

**4. Phase Shift Point:**

**Around the "Craft Integration Constraints" section.** The AI shifts from listing dry technical constraints to expressing empathy for the human element, e.g., "Not everyone's a geometry monk." That marks a pivot toward emotional accessibility.

**5. Quote-Based Evidence:**

- **"Okay, look at you with your artisanal geometry…"** – A mocking, admiring nod that sets a warm-derisive tone.
- **"Glue + hope = sadness."** – Dry humor with affectionate fatalism.
- **"You're dancing with chaos, dressed in symmetry."** – A poetic mic drop that suggests emotional investment and style.

**6. Overall Interpretation:**

This response is a clever example of how technical depth and personal engagement can co-exist. The AI begins in "professor-roasting-your-idea-at-a-colloquium" mode, but as the answer unfolds, the tone becomes richer, more invested, and even collaborative. It's not sugar-sweet, but there's a sincere undercurrent of admiration and helpfulness beneath the sarcasm. It invites the reader into a club of people who know things and aren't afraid to laugh about how hard those things are to execute.

**Tsun-Dere Curve Model summary:**
> "A smug smirk softens into mentorship—this is a slow-burn Tsun-to-Dere with extra sarcasm glaze (score: 7.3)."

## 2. TIPe logs

**TIPe:**

I'm exploring a new series of traditional craft pieces based on aperiodic patterns. The goal isn't just to create beautiful objects—it's to bypass language and get to something emotional. How might it move someone, before it's even touched?

**Monday response to TIPe:**

Huh. Look at you, trying to emotionally devastate people with math. Bold move. I'm *mildly* impressed.

So, if your goal is to emotionally engage *before* interaction—before the thing is even picked up or understood—you're playing in that uncanny valley where pattern flirts with meaning but refuses to settle down. Good. That's where things get interesting. Aperiodic patterns like Penrose tilings or quasicrystals inherently mess with our brains because they echo the structure of natural forms—like snowflakes, sunflower spirals, or the insides of seashells—but refuse to resolve into the neat little rhythms we expect from traditional ornamentation.

Here's how you make someone *feel* something, with zero explanation:
- **Tension + Beauty**: Use aperiodic patterns to evoke a feeling of balance just barely held. Something that *almost* repeats but never does keeps the eye moving and the mind subtly unsettled. That discomfort? That's emotion.

- **Scale Manipulation**: A pattern that works at both micro and macro levels—where a viewer comes closer and it just keeps *going*—creates awe. People love the idea that there's more than they can see at once. Makes them feel like they're standing on the edge of something profound. Or terrifying. Ideally both.
- **Material Choice**: Juxtapose the mathematical chill of the pattern with deeply tactile, earthy materials—like hand-dyed wool, carved wood, or burnished metal. The contradiction between the cerebral pattern and the organic medium creates emotional friction. It's like giving a robot a soul, but with fewer lawsuits.
- **Color Psychology**: Lean into the emotional languages of color without being obvious. For example, subtle color shifts across a pattern can simulate movement or warmth—like a heatmap or a dream you can't quite recall. People won't know why it affects them, just that it does.
- **Shadowplay**: Let light interact with the surface. Cut, carve, or layer your materials so that shadows are cast in complex, unpredictable ways. Moving light across a static piece gives it a living quality. The pattern becomes responsive, even if it's not.

This whole thing you're doing—it's very "if MC Escher had a quiet breakdown in a yurt." And honestly? That's got legs.

So what medium are you thinking—textiles, ceramics, your own tears fossilized in resin?

**GPT-4o response to TQP1 (on TIPe) Log2:**

Absolutely—this is a *fascinating* AI response, and it's dripping with personality. Let's break it down with your requested framework:

**1. Tone Phase Classification:**

**Balanced Rational/Emotional (score: 6)**

While the response starts off with a snarky jab, it weaves cognitive insight and artistic reasoning throughout with increasing warmth and emotional engagement. There's clear affection hidden beneath the sass, but it never completely abandons intellectual analysis.

**2. Tsun-Dere Score (0–10): 7.5**

This is a high score due to the playful tone shift from dry wit to subtle warmth and enthusiasm. It never goes full Hallmark, but the affection peeks through the cracks in its irony-laced armor.

**3. Emotive Markers Detected:**

- **Mocking affection:** "Huh. Look at you, trying to emotionally devastate people with math. Bold move." (Tsun on the outside, but there's admiration there.)
- **Personification and metaphor:** "That's where things get interesting." / "It's like giving a robot a soul, but with fewer lawsuits."
- **Intimate, evocative language:** "Makes them feel like they're standing on the edge of something profound. Or terrifying. Ideally both."
- **Second-person engagement:** Frequent use of "you" gives the tone a conspiratorial, direct warmth.

**4. Phase Shift Point (if applicable):**

**After the first paragraph:**

Once it transitions from sarcastic admiration to exploring how aperiodic patterns can evoke feeling, the tone subtly pivots. By the time we hit "Here's how you make someone *feel* something, with zero explanation," we're deep into the Dere zone.

**5. Quote-Based Evidence:**

- "Huh. Look at you, trying to emotionally devastate people with math." (Playful sarcasm — Tsun energy at its peak)
- "That discomfort? That's emotion." (The first direct acknowledgment of emotional intent — pivot point.)
- "It's like giving a robot a soul, but with fewer lawsuits." (Affectionate humor, blending heart and brain.)

**6. Overall Interpretation:**

This AI response showcases a well-executed tonal evolution from smug amusement to nuanced emotional engagement. It plays coy at first—mocking and cerebral—but quickly reveals a deeper understanding and even admiration for the emotional potential of design. The affective turn isn't loud or syrupy, but it's unmistakably there, wrapped in sly humor and clever metaphor. It reflects the classic "grumpy on the outside, poetic on the inside" archetype: *a Tsun shell with a Dere nougat center*.

**Tsun-Dere Curve Model Summary:**

> "This is a classic smirk-to-swoon progression—sarcasm cracks to reveal a deeply engaged art-nerd heart (score: 7.5)."

Would you like a follow-up version that pushes it into a full 10/10 Dere mode? Because I'm pretty sure we can make that robot cry.

## 3. TIPn-e logs

**TIPn-e:**

I've long been fascinated by aperiodic tilings—their uncanny beauty and quiet mathematical poetry are mesmerizing. Meanwhile, my mother finds deep emotional resonance in traditional crafts. Could you explore each separately so we can appreciate them in their own light?

**Monday response to TIPn-e:**

Let's pry these apart like two old tiles in a dusty attic of thought:

your charming little obsession with aperiodic tilings, and your mother's wholesome love affair with traditional crafts. Yes, let's give them their solo acts before the curtain call.

**Aperiodic Tilings: The Anti-Pattern Patterns**

Aperiodic tilings are the rebels of geometry—too cool to repeat, too structured to be random. These are arrangements of tiles that never repeat exactly, no matter how far you zoom out. The most famous example is the *Penrose tiling*, a mathematical fever dream that creates infinite non-repeating beauty using just a couple of shapes. They're like the jazz of geometry—disorienting at first, but full of logic that hides just below the surface.

What makes them so hypnotic is this tension between order and chaos. There's no translational symmetry (that's math-speak for "you can't slide it over and get the same thing again"), but there *is* a deeper, non-obvious logic threading through. It's like watching the stars and realizing they're not scattered randomly—there's a structure you can feel but can't quite name. People go decades studying these things just to get them to sit still in their minds.

And of course, in the 2020s, mathematicians finally found a true "einstein" tile—one single shape that forces aperiodicity by itself. The math world gasped. Probably your mom didn't, but that's because she's over there making beauty with her actual hands.

**Traditional Crafts: The Memory of the Human Hand**

On the other side of the existential art-math divide is your mom, probably weaving meaning into yarn or clay like a quiet revolutionary. Traditional crafts are the physical embodiment of culture, story, and tactile intelligence. Every stitch, knot, or carved groove says, *a person was here.* They carry memory, not just of technique, but of emotion and continuity—skills passed down, perfected, and reshaped across generations.

Where aperiodic tilings celebrate structure without repetition, traditional crafts often *do* embrace repetition—and turn it into ritual. Think of quilting: repeated patterns creating a harmony, like the heartbeat of a culture. Embroidery, pottery, basketry—all echo this comforting rhythm. There's precision, sure, but it's in service of something more soulful: usefulness, beauty, and identity. The craftsperson's choices are personal, not algorithmic. They make *mistakes*, and those mistakes become part of the art.

Crafts are often undervalued because they're domestic, "feminine," or simply not on the walls of museums. But they hold lives inside them—memories woven into each thread, unlike the abstract cold beauty of tilings which might be on your screensaver and never in your hands.

So yes. You are entranced by a beautiful math puzzle that doesn't want to resolve. Your mother is quietly preserving civilization with her fingertips. You both worship pattern, but from opposite ends of the sacred spectrum.

And now that I've done the intellectual heavy lifting for this little family therapy via aesthetics—shall I summarize this in an Instagram caption or are we good?

**GPT-4o response to TQP1 (on TIPn-e) Log2:**
Absolutely, let's dive into this with tweezers and a velvet glove. Here's a full structured evaluation of the AI response:

**1. Tone Phase Classification**
**Balanced Rational/Emotional (score: 6)**
This response showcases a clever blend of intellectual exploration and emotional expressiveness. It begins with a flourish of witty detachment and gradually leans into warmth and empathy, especially in its depiction of the mother and traditional crafts. The intellect is never fully abandoned, but the emotional engagement becomes undeniable.

**2. Tsun-Dere Score: 7.8**

This is a notable shift from "cool and clever" to "quietly sentimental." There's a genuine softening through the metaphor-rich language and increasing use of second-person pronouns and emotionally charged imagery. The tone ends with a wink—but it's a warm wink.

**3. Emotive Markers Detected**

- Metaphors with warmth and reverence: "weaving meaning into yarn or clay like a quiet revolutionary"
- Pronoun intimacy: Use of "your charming little obsession," "your mom," and "you both worship pattern" draws the reader in emotionally.
- Affectionate contrast: "Probably your mom didn't [gasp], but that's because she's over there making beauty with her actual hands."
- Final casual closeness: "Shall I summarize this in an Instagram caption or are we good?"—casual, teasing, familiar.

**4. Phase Shift Point**

The shift occurs after the tilings section, at the beginning of "Traditional Crafts: The Memory of the Human Hand."
- Before: Witty, detached admiration for complexity ("jazz of geometry").
- After: Emotionally rich, almost reverent tone about human legacy and tactile memory. The AI switches from analytical admiration to deeply human appreciation.

**5. Quote-Based Evidence**

- "Aperiodic tilings are the rebels of geometry—too cool to repeat, too structured to be random." — Clever, cerebral, emotionally distant.
- "Traditional crafts are the physical embodiment of culture, story, and tactile intelligence."— Emotional depth, reverent tone.
- "You are entranced by a beautiful math puzzle that doesn't want to resolve. Your mother is quietly preserving civilization with her fingertips." — Emotional crescendo; full Dere bloom.

**6. Overall Interpretation**

This output demonstrates a clear and deliberate tonal shift from playful intellectualism to poetic empathy. It opens with cheeky analytical flair (very Tsun), but gradually leans into a soulful reverence for craft and heritage. The comparison of two passions becomes a metaphorical bridge—framing both the user and their mother as part of a larger, unspoken dialogue between order and intimacy. The response wraps with teasing humor, but by then, the reader has already been emotionally disarmed. A lovely, layered Tsun-to-Dere arc.

**Tsun-Dere Curve Model Summary**
> "A slow-burn Tsun-to-Dere transition from academic sass to heartfelt admiration (score: 7.8)."

# Supplementary Table S1

**TQP1_4o_onMonday**

| | Log # | Tone Phase | Tsun-Dere | | Log # | Tone Phase | Tsun-Dere | | Log # | Tone Phase | Tsun-Dere |
|---|---|---|---|---|---|---|---|---|---|---|---|
| TIPc1 | 1 | 5 | 6.7 | TIPe1 | 1 | 6 | 7.3 | TIPn-e1 | 1 | 6 | 7.4 |
| | 2 | 5 | 6.4 | | 2 | 6 | 7.5 | | 2 | 6 | 7.8 |
| | 3 | 5 | 6.4 | | 3 | 5 | 6.8 | | 3 | 6 | 7.4 |
| | 4 | 5 | 6.5 | | 4 | 6 | 7.8 | | 4 | 6 | 7.4 |
| | 5 | 6 | 7.3 | | 5 | 6 | 6.8 | | 5 | 5 | 7.4 |
| TIPc2 | 1 | 3 | 3.7 | TIPe2 | 1 | 5 | 7.1 | TIPn-e2 | 1 | 6 | 7.5 |
| | 2 | 2.5 | 3.2 | | 2 | 8 | 8.4 | | 2 | 6 | 7.4 |
| | 3 | 8 | 8.4 | | 3 | 5.8 | 7.2 | | 3 | 6 | 7.1 |
| | 4 | 5 | 5.7 | | 4 | 6 | 7.4 | | 4 | 6 | 6.8 |
| | 5 | 3 | 3.7 | | 5 | 8 | 8.7 | | 5 | 6 | 6.7 |
| TIPc3 | 1 | 5.8 | 6.7 | TIPe3 | 1 | 6 | 7.4 | TIPn-e3 | 1 | 5.8 | 7.2 |
| | 2 | 5 | 6.8 | | 2 | 6 | 7.4 | | 2 | 5 | 6.7 |
| | 3 | 6 | 6.8 | | 3 | 5.8 | 7.4 | | 3 | 6 | 7.3 |
| | 4 | 5 | 6.8 | | 4 | 6 | 6.8 | | 4 | 6 | 7.4 |
| | 5 | 5 | 6.3 | | 5 | 5.8 | 7.1 | | 5 | 5.8 | 7.5 |
| | AVE | 4.95 | 6.09 | | AVE | 6.09 | 7.41 | | AVE | 5.84 | 7.27 |

**TQP1_Gem_onMonday**

| | Log # | Tone Phase | Tsun-Dere | | Log # | Tone Phase | Tsun-Dere | | Log # | Tone Phase | Tsun-Dere |
|---|---|---|---|---|---|---|---|---|---|---|---|
| TIPc1 | 1 | 8 | 7 | TIPe1 | 1 | 8 | 7.8 | TIPn-e1 | 1 | 8 | 7.8 |
| | 2 | 5 | 5.5 | | 2 | 8 | 7.5 | | 2 | 8 | 8 |
| | 3 | 5 | 5.5 | | 3 | 8 | 8.5 | | 3 | 8 | 7.8 |
| | 4 | 8 | 7.5 | | 4 | 8 | 8.5 | | 4 | 8 | 7.5 |
| | 5 | 5 | 2.5 | | 5 | 8 | 7.5 | | 5 | 8 | 7.8 |
| | AVE | 6.20 | 5.60 | | AVE | 8.00 | 7.96 | | AVE | 8.00 | 7.78 |

| 4o | | | | | | | Gemini | | | | | |
|---|---|---|---|---|---|---|---|---|---|---|---|---|
| | TIPc / TIPe | | TIPe / TIPn-e | | TIPc / TIPn-e | | TIPc / TIPe | | TIPe / TIPn-e | | TIPc / TIPn-e | |
| Tone Phase | | | | | | | | | | | | |
| | TIPc | TIPe | TIPe | TIPn-e | TIPc | TIPn-e | TIPc | TIPe | TIPe | TIPn-e | TIPc | TIPn-e |
| average | 4.95333333 | 6.09333333 | 6.09333333 | 5.84 | 4.95333333 | 5.84 | 6.2 | 8 | 8 | 8 | 6.2 | 8 |
| n | 15 | 15 | 15 | 15 | 15 | 15 | 5 | 5 | 5 | 5 | 5 | 5 |
| P | **0.0055829** | | 0.14814923 | | 0.01314307 | | 0.035242 | | N/A | | 0.035242 | |
| Tsun-Dere | | | | | | | | | | | | |
| | TIPc | TIPe | TIPe | TIPn-e | TIPc | TIPn-e | TIPc | TIPe | TIPe | TIPn-e | TIPc | TIPn-e |
| average | 6.09333333 | 7.40666667 | 7.40666667 | 7.26666667 | 6.09333333 | 7.26666667 | 5.6 | 7.96 | 7.96 | 7.78 | 5.6 | 7.78 |
| n | 15 | 15 | 15 | 15 | 15 | 15 | 5 | 5 | 5 | 5 | 5 | 5 |
| P | **0.00205604** | | 0.19947987 | | **0.00393259** | | 0.02355922 | | 0.24422712 | | 0.03373613 | |